\documentclass[manuscript]{acmart}

\AtBeginDocument{%
  }

\copyrightyear{2024}
\acmYear{2024}
\setcopyright{acmlicensed}\acmConference[FAccT '24]{The 2024 ACM Conference on Fairness, Accountability, and Transparency}{June 3--6, 2024}{Rio de Janeiro, Brazil}
\acmBooktitle{The 2024 ACM Conference on Fairness, Accountability, and Transparency (FAccT '24), June 3--6, 2024, Rio de Janeiro, Brazil}
\acmDOI{10.1145/3630106.3659049}

\acmConference[FAccT '24]{The 2024 ACM Conference on Fairness, Accountability, and Transparency}{June 3–6, 2024}{Rio de Janeiro, Brazil}

\acmISBN{979-8-4007-0450-5/24/06}

\usepackage{color, colortbl}
\usepackage{subcaption}

\begin{document}

\title[\textit{Trans}forming Dutch]{\textit{Trans}forming Dutch: Debiasing Dutch Coreference Resolution Systems for Non-binary Pronouns}

\author{Goya van Boven}
\authornote{goya\_jackie@live.nl}
\email{j.g.vanboven@students.uu.nl}
\author{Yupei Du}
\email{y.du@uu.nl}
\author{Dong Nguyen}
\email{d.p.nguyen@uu.nl}
\affiliation{%
  \institution{Utrecht University}
  \streetaddress{Heidelberglaan 8}
  \city{Utrecht}
  \country{the Netherlands}
  \postcode{3584 CS}
}

\renewcommand{\shortauthors}{van Boven et al.}

\begin{abstract}
 Gender-neutral pronouns are increasingly being introduced across Western languages. Recent evaluations have however demonstrated that English NLP systems are unable to correctly process gender-neutral pronouns, with the risk of erasing and misgendering non-binary individuals. This paper examines a Dutch coreference resolution system's performance on gender-neutral pronouns, specifically \textit{hen} and \textit{die}. In Dutch, these pronouns were only introduced in 2016, compared to the longstanding existence of singular \textit{they} in English. We additionally compare two debiasing techniques for coreference resolution systems in non-binary contexts: Counterfactual Data Augmentation (CDA) and delexicalisation. 
 Moreover, because pronoun performance can be hard to interpret from a general evaluation metric like \textsc{lea},  we introduce an innovative evaluation metric, the \textit{pronoun score}, which directly represents the portion of correctly processed pronouns. Our results reveal diminished performance on gender-neutral pronouns compared to gendered counterparts. Nevertheless, although delexicalisation fails to yield improvements, CDA substantially reduces the performance gap between gendered and gender-neutral pronouns. We further show that CDA remains effective in low-resource settings, in which a limited set of debiasing documents is used. This efficacy extends to previously unseen neopronouns, which are currently infrequently used but may gain popularity in the future, underscoring the viability of effective debiasing with minimal resources and low computational costs.
\end{abstract}




\maketitle

\section{Introduction}
Gender-neutral pronouns are increasingly introduced and popularised across Western languages, as suitable alternatives to traditional gendered pronouns for non-binary individuals. The Swedish gender-neutral pronoun \textit{hen} was politically introduced in 2013 \cite{gustafsson2015introducing}, the Dutch \textit{hen/die} was democratically chosen by the Transgender community in 2016 \cite{TNN-2016} and while English has long known singular \textit{they}, the set of \textit{neopronouns} such as \textit{ze} and \textit{thon} is continuously growing \cite{lauscher-etal-2022-welcome}. However, the majority of work in natural language processing (NLP) considers gender as binary and immutable  \cite{cao-daume-iii-2021-toward, devinney2022theories}, thereby excluding transgender individuals, who do not identify with the gender they were assigned at birth. The term \textit{transgender} includes both people with a binary transgender identity (such as transgender women) and people with a non-binary transgender identity. \textit{Non-binary} individuals do not conform to the traditional Western binary categorisation of male or female, identifying for instance as both female and male, as neither or their gender might fluctuate \cite{rajunov2019nonbinary}. 

Within Western societies, transgender people face various forms of discrimination and marginalisation. 
They experience high rates of unemployment, homelessness, 
abuse and poverty,
and frequently experience bullying and discrimination at the workplace \cite{TNNwork, USTransSurvey}. 
Furthermore, transgender people often encounter significant barriers in accessing essential institutions, such as healthcare services and the legal system \cite{zimman2018transgender}. \citet{dev-etal-2021-harms} point out how NLP models can contribute to the marginalisation of transgender people by perpetuating trans-exclusive practices, highlighting the dangers of \textit{erasing} and \textit{misgendering} non-binary individuals. 
 Erasure can occur within NLP, for example, when a system predicts a user's gender but assumes a cisgender identity. Misgendering refers to addressing an individual with a gendered term that does not match their gender identity, which is often experienced as a harmful act \cite{ansara2014methodologies}. 

Recent works have started to investigate non-binary gender biases in NLP systems. In this work we adopt the definition of \textit{bias} provided by \citeauthor{friedman1996bias}: ``computer systems that \textit{systematically} and \textit{unfairly discriminate} against certain individuals or groups of individuals in favor of others'' \cite{friedman1996bias}. Non-binary gender bias evaluations of NLP systems consider 
large language models~\cite{dev-etal-2021-harms, brandl-etal-2022-conservative, watson2023social, hossain-etal-2023-misgendered, martinkova2023measuring, ovalle2023im},
machine translation
\cite{cho-etal-2019-measuring, ghosh2023chatgpt,lauscher-etal-2023-em}, 
POS-tagging \cite{bjorklund2023computer}, and coreference resolution \cite{baumler-rudinger-2022-recognition, cao-daume-iii-2021-toward, brandl-etal-2022-conservative}. In this study we focus on evaluating and debiasing a Dutch coreference resolution system in processing gender-neutral pronouns. Coreference resolution is the task of identifying expressions that refer to the same entity. We focus on this fundamental NLP task, because
any structural mistakes for non-binary individuals at the coreference resolution level, such as failing to recognise their pronouns --- and thereby failing to extract information about these individuals --- can lead to their erasure in downstream applications. 

While earlier studies have evaluated the performance of English coreference resolution systems on gender-neutral pronouns \cite{baumler-rudinger-2022-recognition, cao-daume-iii-2021-toward}, we are the first to perform such an evaluation for a Dutch system, zooming in on the pronouns \textit{hen} and \textit{die}. The Dutch context differs from the English one because (a) Dutch gender-neutral pronouns are less frequent than English singular \textit{they}; (b) in Dutch many nouns are gender-specific, lacking gender-neutral alternatives (e.g. no term like \textit{sibling} exists); and 
(c) there are generally fewer NLP resources available for Dutch than for English. 


We make the following contributions: 1) We systematically evaluate a Dutch coreference resolution system on its performance on gender-neutral pronouns. Our results show that the model currently performs worse on gender-neutral pronouns, compared to gendered counterparts. 2) We  address the limitations of existing evaluation metrics, as these do not provide sufficient insight into the handling of pronouns. To better evaluate systems we propose a new metric, called the \textit{pronoun score}, which directly reflects the percentage of correctly resolved pronouns. 3) We experiment with two different debiasing methods, and find that while delexicalisation~ \cite{lauscher-etal-2022-welcome} does not improve the performance on gender-neutral pronouns in our setup, Counterfactual Data Augmentation does substantially improve the performance on these pronouns. Importantly, our successful debiasing results extend to (a)~ previously unseen neopronouns, and (b) low-resource conditions, which use just a handful of debiasing documents.
 
This paper is structured as follows:  we provide a theoretical background in Section \ref{sec:background} and describe related work in Section \ref{sec:rw}. We continue to describe the data (Section \ref{sec:data}), model (Section \ref{sec:model}) and pronoun score (Section \ref{sec:ps}). We present our experiments in Section \ref{sec:experiments} and finally provide a discussion in Section \ref{sec:discussion}. 
The code used for this project can be found at 
\url{https://github.com/gvanboven/Transforming_Dutch}.

\section{Background}
\label{sec:background}

\subsection{Gender in Dutch}
\paragraph*{Traditional gender manifestation}
In Dutch, gender is expressed in nouns and third-person singular pronouns. There is no gendered verb agreement or case inflection. Table \ref{tab:Dutch_pronouns} summarises the traditional third-person pronouns for animate entities: singular pronouns distinguish between feminine and masculine while there is no gender distinction in plural.
Nouns referring to occupations and family members are usually gendered. For many occupations, the masculine form is the root (e.g. \textit{eigenaar} (\textit{owner}), \textit{schrijver} (\textit{writer})) and the feminine form adds a suffix (e.g. \textit{eigenares}, \textit{schrijfster}) \cite{gerritsen2002language}. In other cases the male form is used for all genders (e.g. \textit{professor} (\textit{professor})). For some occupational terms gender-neutral alternatives exist, e.g. replacing \textit{lerares} (\textit{female teacher}) and \textit{leraar} (\textit{male teacher}) with \textit{leerkracht} (\textit{teacher}), but for many occupations this is not the case. Similarly, most words describing relatives only have a feminine and masculine version: e.g. no term like \textit{cousin} exists in Dutch, only providing the options of male \textit{neef} 
and female \textit{nicht}. 
This is a problem for non-binary people, since there is no alternative that matches their gender identity. 

\begin{table}[]
\centering
\small
    \caption{Overview of the traditional Dutch third-person pronouns for animate entities. The neuter singular form \textit{het} is excluded because this form is only used for inanimate entities. }
    \label{tab:Dutch_pronouns}
\begin{tabular}{llll}
    \toprule
   &  \multicolumn{2}{c}{\textbf{Singular}} & \textbf{Plural} \\
    Gender    & Feminine & Masculine & All genders \\
    \midrule
    Personal (subject) & \textit{zij} & \textit{hij}   & \textit{zij}\\
    Personal (direct object) & \textit{haar} & \textit{hem} & \textit{hen}\\
    Possessive & \textit{haar} & \textit{zijn} & \textit{hun} \\
    \bottomrule
\end{tabular}
\end{table}

\paragraph*{Gender-neutral pronouns}
In 2016, Transgender Netwerk Nederland organised a vote to determine the Dutch gender-neutral pronouns, in which 500 community members participated. Here, \textit{hen/hen/hun} was favoured over \textit{die/die/diens} and neopronouns \textit{dee/dem/dijr} \cite{TNN-2016}. However, as of 2024, both \textit{hen} and \textit{die} are increasingly being adopted by non-binary people \cite{rtlnieuws-2021, trouw-2020}. Additionally, a broader set of neopronouns has been proposed, including \textit{zhij} and \textit{ij} \cite{VK-Hurkens-2021, PronounsPage}, but these are not as widely used yet. 
Traditionally, \textit{die} is a demonstrative and relative pronoun, and \textit{hen} is a third-person plural personal pronoun (i) for direct objects and (ii) succeeding prepositions. When used as gender-neutral pronouns, there is no difference in meaning between the two, and they can be used interchangeably. Contrasting the English singular \textit{they} that remains conjugated as plural, both gender-neutral \textit{hen} and \textit{die} are conjugated as singular. An example usage of \textit{hen} and \textit{die} in Dutch is: ``Noa geeft \textit{hun/diens} studieboeken weg omdat \textit{hen/die} \textbf{is} afgestudeerd'' (Noa gives \textit{their} study books away because \textit{they} \textbf{have} graduated);
note that singular \textbf{is} is used in Dutch while plural \textbf{have} is used in English. 

\subsection{Coreference resolution}
Coreference resolution entails deciding whether two referring expressions \textit{corefer}, i.e. whether they refer to the same entity. \textit{Referring expressions} or \textit{mentions} are linguistic expressions that are used to refer to entities. 
A \textit{cluster} is a set of coreferring expressions. An entity that only has a single mention is called a \textit{singleton}. Within a cluster, the mentions that precede a certain mention are called its \textit{antecedents} while its later mentions are \textit{anaphors} or \textit{anaphoric}. The following sentence 
shows an example, in which mentions are marked in brackets, and mentions with the same colour refer to the same entity: ``[\colorbox{pink!50}{Sam Smith}] is a famous singer. [\colorbox{pink!50}{They}] collaborated with [\colorbox{blue!30}{Kim Petras}] recently.''
Coreference resolution consists of two subtasks, which end-to-end systems perform simultaneously~\cite{lee-etal-2017-end, lee-etal-2018-higher}: (1) mention detection, i.e. identifying the spans of referring expressions and (2) identifying the coreference links between the mentions. 

\section{Related work}
\label{sec:rw}
\paragraph*{Non-binary gender bias evaluations}
Recent works evaluate non-binary gender bias in English language models~\cite{dev-etal-2021-harms, hossain-etal-2023-misgendered, watson2023social, ovalle2023im}, and in multi-lingual evaluations~\cite{brandl-etal-2022-conservative, martinkova2023measuring}. In downstream tasks, non-binary gender bias evaluations are performed for machine translation \cite{lauscher-etal-2023-em}, NER \cite{lassen2023detecting}, POS-tagging \cite{bjorklund2023computer} and abusive language detection \cite{sobhani2023measuring} systems. In coreference resolution, the WinoNB dataset \cite{baumler-rudinger-2022-recognition} is designed 
to test systems' ability to disambiguate singular and plural \textit{they}. Moreover, the GICoref dataset \cite{cao-daume-iii-2020-toward} 
contains naturally occurring data of ``gender-related phenomena''\cite{cao-daume-iii-2020-toward}, and 
a relatively balanced distribution of \textit{he}, \textit{she}, \textit{they} and neopronouns. Both datasets reveal poor model performances on gender-neutral pronouns. 
Finally, \citet{brandl-etal-2022-conservative} evaluate a Danish coreference resolution model on gender-neutral pronouns. They create a gender-neutral version of a regular coreference resolution corpus by replacing gendered pronouns with gender-neutral ones. They report a small CoNLL score performance drop on the gender-neutral data. However, from their results it does not become entirely clear \textit{how many} more gender-neutral pronouns are incorrectly resolved, as the CoNLL-score does not directly reflect this.


\paragraph*{Debiasing}
To our best knowledge, only two studies consider debiasing NLP systems for non-binary gender bias. First, \citet{hossain-etal-2023-misgendered} aim to improve language model performance in incorporating the declared preferred pronouns of an individual. They explore few-shot in-context learning using explicit examples and note an improvement in performance. However, the improvement plateaus rapidly, falling short of achieving comparable accuracy levels to those observed for gendered pronouns. Second, \citet{bjorklund2023computer} aim to improve the POS-tagging performance on the Swedish gender-neutral pronoun \textit{hen}. They first augment the training data with semi-synthetic data by replacing gendered pronouns with gender-neutral ones. Then, they fine-tune their models from scratch on the augmented data. Encouragingly, they observe that including the gender-neutral pronoun in 2\% of the training sentences is sufficient to remove the performance gap. We consider their study the most similar to the current work as (a) both works involve debiasing downstream tasks and (b) their debiasing method is similar to our application of CDA. However, besides considering a different task and language, we make the additional contributions of (1) evaluating a ``continual fine-tuning'' debiasing configuration, besides fine-tuning from scratch; (2)~ additionally evaluating the delexicalisation method; and (3) investigating the effect of the debiasing methods on previously unseen pronouns.

\section{Data}
\label{sec:data}
In this section, we introduce and analyse the data (Section~\ref{sec:data_analysis}); describe the data preprocessing steps (Section \ref{sec:data_preprocessing}); and finally describe the transformation steps for inserting gender-neutral pronouns into the corpus (Section~\ref{sec:data_transformation}).

\subsection{Data analysis}
\label{sec:data_analysis}

We use the 1M-token SoNaR-1 corpus \cite{reynaert-etal-2010-balancing,oostdijk2013sonar}, because this is the largest Dutch corpus annotated for coreference resolution to date. SoNaR-1 consists of 861 documents from various domains, e.g. magazines, Wikipedia articles, brochures, websites, legal texts, autocues and press releases. The coreference relations were manually annotated based on the COREA guidelines \cite{hendrickx-etal-2008-coreference}. The corpus also contains manually checked annotations for syntactic dependency trees, spatio-temporal relations, semantic roles and named entities.

Personal and possessive pronouns constitute 2.9\% of all tokens in the corpus. Of these pronouns, third-person singular pronouns are the most prevalent, constituting 59.2\%. A striking gender imbalance can be observed: \textbf{79.1\% of all third-person pronouns are masculine}. Table \ref{tab:sonar_pronoun_frequencies} displays the overall frequency of gender-neutral pronouns and neopronouns. While some of the types frequently appear in the corpus, further analysis of the POS-tags reveals that none of these types are actually used as third-person singular pronouns within the corpus, indicating that \textbf{the corpus does not include gender-neutral pronouns or neopronouns}. 

\begin{table}[]
\centering
\small
\caption{Frequency of gender-neutral pronouns and neopronouns in the SoNaR-1 corpus. \textit{Die} appears 1,995 times as a demonstrative pronoun, 5,268 times as a relative pronoun, and 19 times with an alternative label. Its possessive form, \textit{diens}, appears 35 times as a demonstrative pronoun. The 290 occurrences of \textit{hen} function as a third-person plural object personal pronoun. Its possessive counterpart, \textit{hun}, appears 1,865 times as a third-person plural possessive pronoun. \textit{Vij} appears once and \textit{zeer} (also connoting \textit{very} or \textit{sore}) occurs 295 times as an adverb. The remaining neopronouns do not feature in the dataset. None of the considered gender-neutral pronouns and neopronouns appear as third-person singular pronouns in the corpus.}
\label{tab:sonar_pronoun_frequencies}
\begin{tabular}{lrlr}
\toprule
\textbf{Pronoun} & \textbf{Frequency} & \textbf{Pronoun}     & \textbf{Frequency}  \\
\toprule
\textit{die}     &  7,282    & 
\textit{zeer}    &  295        \\
\textit{diens}   &  35         & 
\textit{vij}     &  1         \\
\textit{hen}     &  290        & 
\textit{dee, dem, dijr, dij, dem,}   &  \\
\textit{hun}     & 1,865     &
\textit{dijr, nij, ner, nijr, vijn}      &   0    \\
& & \textit{vijns, zhij, zhaar, zem} & \\
\bottomrule
\end{tabular}
\end{table}

\subsection{Data preprocessing}
\label{sec:data_preprocessing}

We use the genre-balanced 70/15/15 division by \citet{poot-van-cranenburgh-2020-benchmark} to divide the documents over train/dev/test splits. 
Moreover, we remove all singleton annotations from SoNaR-1. The reason for this is two-fold: (1)~ While the SoNaR-1 corpus includes singleton annotations, coreference resolution systems are frequently trained and evaluated on corpora without singleton annotations \cite{lee-etal-2018-higher, joshi-etal-2020-spanbert, liu-etal-2022-autoregressive, bohnet-etal-2023-coreference}; (2) Pronouns typically refer to proper nouns and thus seldom exist as singletons. Recognising singletons thus appears to be of limited relevance for the current study.


\subsection{Construction of pronoun-specific data}
\label{sec:data_transformation}

To compare the performance on different pronouns, we create four \textit{pronoun-specific} versions of the test split, in which all third-person pronouns are replaced by a specific pronoun set: 1. \textit{hij/hem/zijn} (masculine), 2. \textit{zij/haar/haar} (feminine), 3. \textit{hen/hen/hun} (gender-neutral) and 4. \textit{die/hen/diens} (gender-neutral). Examples can be found in Table \ref{tab:rewrite_examples}. We use a rule-based rewriting algorithm based on \citet{zhao-etal-2018-gender}. The algorithm consists of three steps. The principle step is swapping pronouns. Moreover, 
names are anonymised and gendered nouns are replaced by gender-neutral nouns. These operations ensure that the performance on the different \textit{pronoun-specific} test sets can be fairly compared, as they avoid confounding gender effects in case the gender associations of a name or noun and a pronoun no longer match after pronoun swapping, as for example in the phrase \textit{Vader en haar kind} (\textit{Father and her child}).



\begin{table}[t]
\small
\centering
\caption{Example sentence in the SoNaR-1 dataset, before and after transforming it into different pronoun settings. Words that are changed between the versions are marked in bold. Here, the noun replacement \textit{vrouw} (\textit{wife} in this context, but also means \textit{woman})$\rightarrow$\textit{persoon (person)} changes the meaning of the noun in this context. We consider this replacement incorrect. We manually evaluated a subset of the replacements, and only 1.53\% of the replacements was incorrect.}
\label{tab:rewrite_examples}
\begin{tabular}{lll}
\toprule
\textbf{Dataset}      & \textbf{Gender}    & \textbf{Sentence}     \\
\midrule
\textit{Original}     &  Masculine  & \begin{tabular}[c]{@{}l@{}}Na \textbf{zijn} herstel vindt \textbf{hij} \textbf{zijn} \textbf{vrouw} en \textbf{zijn} \textbf{moeder} terug in Folkestone. \\   \textit{After \textbf{his} recovery \textbf{he} finds \textbf{his} \textbf{wife} and \textbf{his} \textbf{mother} back in Folkestone.}     \end{tabular}                       
\\
\rowcolor[HTML]{EFEFEF} 
\textit{Pronoun-specific hij} & Masculine &  \begin{tabular}[c]{@{}l@{}}
 Na \textbf{zijn} herstel vindt \textbf{hij} \textbf{zijn} \textbf{persoon} en \textbf{zijn} \textbf{ouder} terug in Folkestone. \\
 \textit{After \textbf{his} recovery \textbf{he} finds \textbf{his} \textbf{person} and \textbf{his} \textbf{parent} back in Folkestone.}\\
     \end{tabular}                                                                                                       \\
\textit{Pronoun-specific zij} & Feminine & Na \textbf{haar} herstel vindt \textbf{zij} \textbf{haar} \textbf{persoon} en \textbf{haar} \textbf{ouder} terug in Folkestone. \\ 
\rowcolor[HTML]{EFEFEF} 
\textit{Pronoun-specific hen} & Gender-neutral & Na \textbf{hun} herstel vindt \textbf{hen} \textbf{hun} \textbf{persoon} en \textbf{hun} \textbf{ouder} terug in Folkestone. \\
\textit{Pronoun-specific die} & Gender-neutral & Na \textbf{diens} herstel vindt \textbf{die} \textbf{diens} \textbf{persoon} en \textbf{diens} \textbf{ouder} terug in Folkestone. \\  
\bottomrule  

\end{tabular}
\end{table}

\textbf{1. Swapping pronouns~} We recognise third-person singular pronouns by their POS-tag\footnote{This is necessary because some Dutch pronouns have identical lexical forms but distinct grammatical functions, such as possessive or personal object \textit{haar} and third-person singular or plural \textit{zij}.}, and replace them according to the rules stipulated for the targeted dataset version (e.g., \textit{hij} $\rightarrow$ \textit{hen} for the \textit{pronoun-specific} \textit{hen} test set). 

\textbf{2. Name anonymisation~} Following \citeauthor{zhao-etal-2018-gender}, we recognise names using named entity annotations,\footnote{ We use the  manually checked named entity annotations included in the SoNaR-1 corpus for this.} considering all tokens with a \texttt{PER} (person) tag. 
Names are replaced with a standardised tag \texttt{ANON\_\textit{x}}, with  $x\in \mathbb{N}$, where the same value of $x$ always replaces the same string. For example:\footnote{This step may introduce some processing difficulties to coreference models, because these models are usually not trained on data with anonymised names. But, as this step is performed across all data versions, it will affect all test sets in the same way.}
\begin{small}
    \begin{center}
    \begin{tabular}{ccc}
         \textit{Jan Jansen is op vrijdag vrij omdat Jan dan voetbalt}& $\rightarrow$ & \texttt{ANON\_0} \texttt{ANON\_1} \textit{is op vrijdag vrij omdat} \texttt{ANON\_0} \textit{dan voetbalt} \\
         (\textit{Jan Jansen is free on Friday because Jan plays football})& $\rightarrow$ & ( \texttt{ANON\_0} \texttt{ANON\_1} \textit{is free on Friday because} \texttt{ANON\_0} \textit{plays football}.)
    \end{tabular}    
\end{center}
\end{small}

\textbf{3. Replacing gendered nouns~}  We create a Dutch list (Appendix \ref{sec:app_rewriting_rules}) of gendered nouns and their gender-neutral counterparts (e.g. \textit{moeder} (\textit{mother}) $\rightarrow$ \textit{ouder} (\textit{parent})), using the English list by \citet{zhao-etal-2018-gender} as a basis. To ensure the quality of replacements, a panel of six individuals participated in reviewing the list and contributed their suggestions.\footnote{Among these individuals, three use \textit{she/her} pronouns, two use \textit{he/him} pronouns, and one uses \textit{he/they} pronouns. The recruitment of participants was facilitated through our personal network.} 

\paragraph*{Limitations} For some nouns it proved challenging to identify gender-neutral replacements, as numerous Dutch words exclusively have gendered forms (e.g. \textit{nicht} (\textit{niece}) lacks an alternative akin to \textit{cousin}).\footnote{In such instances, we settled for a gender-neutral hypernym of the term of interest, such as \textit{familielid} (\textit{family member}), sacrificing a substantial portion of the meaning of the original term. Such instances are marked in Appendix \ref{sec:app_rewriting_rules}.} Other gendered terms have multiple meanings (e.g. \textit{vrouw} means both \textit{woman} and \textit{wife}), forcing the selection of a one interpretation for the replacement, as illustrated in Table \ref{tab:rewrite_examples}. Moreover, sometimes changing a noun in Dutch necessitates changing the determiner (e.g. \textit{de dochter} (\textit{the daughter}) $\rightarrow$ \textit{het kind} (\textit{the child})), but we did not extend the transformation algorithm to feature this ability. We manually review a subset comprising 12,584 tokens, balanced over the data versions and splits. Out of 1,111 replacements,
only 17 are incorrect, giving an error rate of $1.53\%$. We therefore consider the transformed data to be of good quality.

\section{Model}
\label{sec:model}

In alignment with prior debiasing studies, which only consider debiasing neural coreference resolution systems \cite[e.g.][]{zhao-etal-2018-gender}, we focus on debiasing a neural Dutch model.\footnote{We therefore exclude the rule-based Dutch dutchcoref \cite{van2019dutch} model and the hybrid Dutch model by \citet{van-cranenburgh-etal-2021-hybrid}.}
We select the wl-coref \cite{dobrovolskii-2021-word} model, which is originally trained on English data. We chose this model because (a) it achieves a competitive performance, (b) it has a low complexity and (c) its base models are available in Dutch.


\subsection{Architecture}
\label{sec:model_architecture}

The wl-coref model \cite{dobrovolskii-2021-word} architecture consecutively performs the following two steps:
(1) Predicting the antecedent for each word individually, or predicting that the word does not have an antecedent. During this phase, the model exclusively focuses on identifying antecedents for the \textit{heads} of mention spans. The span's head is defined as the only word in the span with a head outside of the span, or as the root of the sentence. For example,  the head of the mention \textit{their roommate} in Figure \ref{fig:pronoun_score_example} is \textit{roommate}, as the head of this word, \textit{asked}, is outside of the mention span.
(2) Predicting the full mention span boundaries from the mention heads, ultimately culminating in the final coreference predictions for complete mentions.


\subsection{Training}
\label{sec:model_training}
We fine-tune the wl-coref architecture on the SoNaR-1 corpus, without making any changes to the core modules. Following \citet{dobrovolskii-2021-word}, we train the models for 20 epochs and report the performance of the epoch that performs best on the development data. We make three minor adjustments to the setup: 
\begin{enumerate}
    \item We change the evaluation metric from the \textsc{CoNLL} score to the \textsc{lea} score, as the \textsc{CoNLL} metric has been demonstrated to be flawed \cite{moosavi-strube-2016-coreference}. For the wl-coref model, the performance scores can be computed both at the word-level (before span boundary prediction) and at the span-level (after span boundary prediction). We report the span-level performance, because this represents the performance on the full coreference resolution task.
    \item While wl-coref uses speaker and genre information, 
    the SoNaR-1 corpus does not contain speaker information. Therefore, we use the same speaker value (zero) for all instances. We also experiment with taking out the speaker component entirely, but this does not improve the performance.
    \item Although genre information is available for SoNaR-1, we follow \citet{poot-van-cranenburgh-2020-benchmark} in always using the same genre value. We leave exploring the effect of including genre information to future work. 
\end{enumerate} 

We use the Hugging Face Transformers \cite{wolf-etal-2020-transformers} implementations to compare three base models: the Dutch 
robBERT model \cite{delobelle2020robbert}, and the multilingual mBERT \cite{devlin-etal-2019-bert} and XLM-RoBERTa \cite{conneau-etal-2020-unsupervised} models, all in their \textit{base} versions. 
Table \ref{tab:base-models} shows the results on the SoNaR-1 dev set, using the same hyperparameters as \citet{dobrovolskii-2021-word}. As XLM-RoBERTa obtains the best performance (F1-score = 52.4), we use this model in our main experiments. Furthermore, we execute a hyperparameter search (full results in Appendix \ref{app:hyperparams}) for two hyperparameters: the learning rate (best value = $5e^{-4}$) and the BERT learning rate (best value = $3e^{-5}$). We also experiment with lowering $k$ and using a different learning rate schedule than the standard linear one, but these adaptations do not improve the performance.

We report the final performance scores in Table \ref{tab:modeltestresults}, as the average of five random seeds. The model obtains a test F1-score of 55.6. For a comparison between wl-coref and other Dutch coreference resolution models, refer to Appendix~\ref{app:models}. Throughout the rest of this study, we report the main results as the average of five random seeds. 


\begin{figure}[t]
    \centering
    \begin{minipage}[t]{0.5\textwidth}
        \centering
        \small
        \captionof{table}{\textsc{lea} performances of the wl-coref model on the development set of the SoNaR-1 corpus using three different base models.}
        \label{tab:base-models}
        \begin{tabular}{lcc}
        \toprule
        \textbf{Base model}                                & \multicolumn{1}{l}{\textbf{Dev F1}} \\
        \midrule
        \rowcolor[HTML]{EFEFEF} 
        robBERT \cite{delobelle2020robbert}          & 45.5      \\
        mBERT-base \cite{devlin-etal-2019-bert}      & 47.0       \\
        \rowcolor[HTML]{EFEFEF} 
        XLM-RoBERTa-base \cite{conneau-etal-2020-unsupervised} & \textbf{52.4 }                \\
        \bottomrule
        \end{tabular}

    \end{minipage} \hspace{0.2cm}%
    \begin{minipage}[t]{0.45\textwidth}
        \centering
        \small
         \captionof{table}{Average \textsc{lea} performance scores of the wl-coref model with XLM-RoBERTa-base on the SoNaR-1 development and test set, using five random seeds.}
        \label{tab:modeltestresults}
        \begin{tabular}{lccc}
        \toprule
        \textbf{Data} & \multicolumn{1}{c}{\textbf{Precision}} & \multicolumn{1}{c}{\textbf{Recall}} & \multicolumn{1}{c}{\textbf{F1}} \\
        \midrule
        Dev   &  53.0 ($\sigma = 2.3$) &	57.9 ($\sigma = 3.1$) &	55.3 ($\sigma = 0.2$)
                                  \\
        Test  &    55.5 ($\sigma = 2.3$)& 	55.8 ($\sigma = 2.7$)&  	55.6 ($\sigma = 0.5$) \\
        \bottomrule
        \end{tabular}

    \end{minipage}
\end{figure}

\begin{figure}[t]
\centering
\small
\begin{subfigure}[b]{0.9\textwidth}
   [\colorbox{pink!50}{Raven}] entered the kitchen. ``Did [\colorbox{cyan!30}{you}] sleep well?'', [\colorbox{pink!50}{they}] asked [\colorbox{cyan!30}{}[\colorbox{pink!50}{their}] \colorbox{cyan!30}{roommate}]. ``No [\colorbox{pink!50}{Raven}]'', said [\colorbox{cyan!30}{Thorn}] annoyed, ``[\colorbox{blue!30}{Tobi}] called me way too early.''
   \caption{}
\label{fig:pronoun_score_example_gold} 
\end{subfigure}

\begin{subfigure}[b]{0.9\textwidth}
   [\colorbox{pink!50}{Raven}] entered the kitchen. ``Did [\colorbox{cyan!30}{you}] sleep well?'', they asked [\colorbox{cyan!30}{}[\colorbox{pink!50}{their}] \colorbox{cyan!30}{roommate}]. ``No
   [\colorbox{pink!50}{Raven}]'', said [\colorbox{cyan!30}{Thorn}] annoyed, ``[\colorbox{blue!30}{Tobi}] called me way too early.''
   \caption{}
   \label{fig:pronoun_score_example_predicted}
\end{subfigure}
\caption[Example annotations and predictions]{An example sentence, with its gold annotations in (a), and example predictions in (b). Mentions are indicated with brackets. Mentions with the same colour belong to the same cluster.}
\label{fig:pronoun_score_example}
\end{figure}

\section{Pronoun score}
\label{sec:ps}

We use link-based entity aware \textsc{lea} \cite{moosavi-strube-2016-coreference} as our evaluation metric. Because 
the \textit{pronoun-specific} test sets 
only differ in third-person pronouns, any performance difference between these sets can be attributed to third-person pronouns. However, if the \textsc{lea} F1-score is e.g. one point lower for \textit{hen} than for \textit{hij} pronouns, it is not directly clear how many more \textit{hen} pronouns are incorrectly resolved. Per illustration, Figure \ref{fig:pronoun_score_example_gold} shows a sentence with gold annotations, and \ref{fig:pronoun_score_example_predicted} shows an example prediction. The prediction is correct, except the pronoun \texttt{[they]} is not recognised  
as a mention, and is thus not considered as part of the \textit{Raven} cluster. This prediction results in a \textsc{lea} F1-score of $\frac{6}{7}$.\footnote{For an explanation of how the \textsc{lea} score is computed, refer to \citet{moosavi-strube-2016-coreference}} From this score alone, it is hard to interpret that $\frac{1}{2}$ of the third-person pronouns is correctly resolved. To get a more direct insight into the model's ability to process pronouns, we introduce the \textbf{pronoun score}, complementary to the \textsc{lea} score, that represents \textit{the percentage of pronouns for which at least one correct antecedent is identified}. As we focus on third-person singular pronouns in this work, we only consider pronouns in this category, and define the metric as:
\begin{displaymath} pronoun\_score = \frac{
\sum_{p\in third\_person\_pronouns} [ ( gold\_ants(p) \cap predicted\_ants(p) >= 1 ]}{|pronouns|} \cdot 100 \% \end{displaymath}

\noindent We illustrate this score with the example in Figure \ref{fig:pronoun_score_example}. We can identify the following gold and predicted antecedents for the third-person pronouns \textit{they} and \textit{their}: 
\begin{center}
\small
\begin{tabular}{llll}
 $gold\_ants (they)$    & $ = \{ Raven \}$ &  $predicted\_ants (they)$ & $ = \{ \}$ \\
 $gold\_ants (their)$    &$ = \{they, Raven \}$ &  $predicted\_ants (their)$ & $ =\{Raven\}$ \\
\end{tabular}
\end{center}

\noindent Then the pronoun score is computed as follows: 
\begin{center}
\small
\begin{tabular}{lll}
$gold\_ants(they) \cap predicted\_ants(they)>=1$ &  $= \{ Raven \} \cap \{ \}>=1$ &$= 0 >= 1 =0 $ \\
$gold\_ants(their) \cap predicted\_ants(their)>=1 $ & $ = \{they, Raven \} \cap \{Raven\}>=1$ &$= 1>=1=1$ \\
\end{tabular}
\[pronoun\_score= \frac{0+1}{2}\cdot 100 \% = \frac{1}{2}\cdot 100 \% = 50\%\]
\end{center}

\noindent The pronoun score thus directly reflects the amount of correctly resolved third-person pronouns.

\paragraph*{Design considerations} Rather than only considering the one antecedent that is \textit{directly} predicted by the model, we consider \textit{all} antecedents in the predicted cluster in the evaluation, and consider at least one correct antecedent in the prediction cluster to be sufficient. We prefer this more relaxed configuration because prior evaluations show poor performances on gender-neutral pronouns \cite{baumler-rudinger-2022-recognition, cao-daume-iii-2021-toward}, indicating the difficulty of the task. However, the metric is highly adaptable, and can be made stricter for more straightforward tasks. Moreover, we prefer considering one correct antecedent as sufficient, over an alternative such as requiring all pronoun antecedents to be correct, because otherwise the model might be punished twice for a single mistake.\footnote{This is illustrated by the example in Figure \ref{fig:pronoun_score_example}. In the predictions, \textit{they} is missed as a mention, and therefore \textit{their} also misses one of its correct antecedents. Requiring all the correct antecedent to be found results in a pronoun score of zero for this example, punishing the mistake for \textit{they} twice. We consider this outcome undesirable.}  The \textsc{lea} score already provides a holistic view of the model's performance, so we prefer the pronoun score to complement the \textsc{lea} score by zooming in on the pronoun alone.\footnote{For the same reason, we decide not to incorporate other error  types (e.g., false positives) in the pronoun score: this would make the metric less interpretable, and the \textsc{lea} score already has a recall and precision component.} 

A potential objection against considering antecedents is that the first mention of a cluster is always excluded from the evaluation. However, pronouns are typically used to replace names or proper nouns, and thus rarely appear as the first mention of a cluster. This objection thus does not appear to be relevant for pronouns.

\section{Experiments}
\label{sec:experiments}
In Section \ref{sec:exp1} we compare between the performance on gendered pronouns and gender-neutral pronouns. In Section \ref{sec:debiasing-experiments} we evaluate the effectiveness of two debiasing techniques: Counterfactual Data Augmentation (CDA) and delexicalisation. Next, in Section \ref{sec:debiasing_data_amount} we explore CDA debiasing in low resource conditions. Finally we evaluate the performance of the original and the debiased models on previously unseen neopronouns in Section \ref{sec:unseen_pronouns_exp}. 

\subsection{Gender-neutral pronoun evaluation experiment}
\label{sec:exp1}

\subsubsection{Setup}
We compare the wl-coref model's performance on gendered and gender-neutral pronouns, by comparing its performance on the \textit{pronoun-specific} versions of the test set (Section \ref{sec:data_transformation}). 
Besides changing pronouns, the transformation performed to create the \textit{pronoun-specific} test sets involves obscuring gender clues through (a) replacing gendered nouns by gender-neutral nouns, and (b) anonymising names. Prior to performing the experiment, we compute how these two transformations affect the model's performance, to isolate the impact of changing the pronouns in the \textit{pronoun-specific} data sets (Table \ref{tab:exp1_results}, \textit{baseline}). We observe a drop of 4.16 points in F1-score compared to the original test data. 

\begin{table}[]
\centering
\small
\caption{Gender-neutral pronoun evaluation experiment: \textsc{lea} and pronoun scores on the \textit{pronoun-specific} test sets, as the average of five random seeds. The \textit{baseline} performance refers to the performance on the version of the test set in which gender clues are removed, but the pronouns remain unchanged. The model's performance is lower on gender-neutral than on gendered pronouns.}
\label{tab:exp1_results}
\begin{tabular}{lcccc|cl}
\toprule
 \textbf{Data}                          & \textbf{Precision} & \textbf{Recall} & \textbf{F1} & \textbf{$\Delta$ F1 baseline} & \textbf{Pronoun score} & \textbf{$\Delta$ with \textit{hij}} \\
\midrule
\textit{Baseline} & 53.58 ($\sigma$=2.24)         & 49.59 ($\sigma$=2.62)           & 51.41 ($\sigma$=0.44)        & -          \\
\hline
\textit{Hij (masculine)}      & 52.23 ($\sigma$=2.30)                      & 49.66  ($\sigma$=2.76)                              & 51.29 ($\sigma$=0.42)                 & -0.12      & 88.36\% ($\sigma=$0.89) & -    \\
\textit{Zij (feminine)}       & 53.18 ($\sigma$=2.33)                                  & 48.73 ($\sigma$=2.56)                              & 50.77 ($\sigma$=0.37)                & -0.64    & 86.65\%  ($\sigma=$1.23)  &   -1.71       \\
\textit{Hen (gender-neutral)} & 53.29 ($\sigma$=2.56)                                  & 45.82 ($\sigma$=3.16)                              & 49.14 ($\sigma$=0.68)                 &  -2.27    & 75.85\%    ($\sigma=$2.93)   &   -12.51  \\
\textit{Die (gender-neutral)} & 52.55 ($\sigma$=1.46)  & 44.94 ($\sigma$=2.24)        & 48.36 ($\sigma$=0.44)                 & -3.05  & 57.49\%  ($\sigma=$6.55)   & -30.87   \\
\bottomrule
\end{tabular}
\end{table}

\subsubsection{Results}
\label{sec:exp1_results}
Table \ref{tab:exp1_results} reports the performances on the \textit{pronoun-specific} test sets. We now discuss the main observations.

\textbf{The best performance is achieved on \textit{hij} pronouns} 
(\mbox{pronoun score = 88.36\%}; \mbox{F1=51.29}). This is according to expectations, as masculine pronouns constitute 79,1\% of the third-person pronouns in the training data (see Section \ref{sec:data_analysis}). 

\textbf{The performances on \textit{hij} and \textit{zij} pronouns are similar} (-0.64 in F1-score and -1.71 percentage points in pronoun score on \textit{zij} compared to \textit{hij} pronouns). This is surprising, because earlier studies found English coreference resolution models to perform better on masculine than on feminine pronouns \cite{webster-etal-2018-mind, kurita-etal-2019-measuring, rudinger-etal-2018-gender}.\footnote{A difference with English is that in Dutch, the feminine third-person singular pronoun \textit{zij} (250 occurrences in the corpus) is also used as a third-person plural pronoun (481 occurrences). This might increase the model's familiarity with the type, and boost it's recognition as a pronoun.} 

\textbf{The performance is worse on gender-neutral pronouns.} 
The pronoun score drops strongly for \textit{hen} pronouns (-12.51 percentage points) and decreases even further for \textit{die} (-30.87 percentage points). 

\textbf{\textit{Hen} pronouns are better resolved than \textit{die} pronouns.} The high standard deviations for \textit{die} additionally indicate an unstable resolution. A potential reason for this is that \textit{hen} is always used as a personal or possessive pronoun in Dutch (be it as a plural pronoun), whereas this is not the case for \textit{die} (see Table \ref{tab:sonar_pronoun_frequencies}).

\subsection{Debiasing experiment}
\label{sec:debiasing-experiments}

\subsubsection{Setup}
In this experiment we compare two debiasing methods to improve the performance on gender-neutral pronouns. The first debiasing method is \textbf{Counterfactual Data Augmentation} (CDA), which involves replacing existing pronouns in the training data with the pronouns of interest \cite{zhao-etal-2018-gender, zhao-etal-2019-gender}. We create a \textit{gender-neutral} version of the data, following the algorithm described in Section \ref{sec:data_transformation}, in which all third-person singular pronouns are substituted by gender-neutral pronouns: \textit{hen} in 50\% of the documents, and \textit{die} in the remaining 50\%. The rationale behind this methodology is that inserting gender-neutral pronouns into the training data is expected to improve the model's processing of these pronouns. Given the favourable outcomes demonstrated by CDA in mitigating binary-gender bias within coreference resolution systems \cite{zhao-etal-2018-gender, zhao-etal-2019-gender}, we expect this method to be effective. 

The second debiasing method is \textbf{delexicalisation} \cite{lauscher-etal-2022-welcome}. This method is applied by training the model on a \textit{delexicalised} version of the data, wherein all third-person pronouns are replaced by their corresponding POS-tag. We adapt \citet{lauscher-etal-2022-welcome}'s approach slightly by replacing the POS-tags, which only distinguish between personal and possessive pronouns, by syntactical tags for subjects (\texttt{<SUBJ>}), objects (\texttt{<OBJ>}) and possessives (\texttt{<POSS>}). This adaptation is made to distinguish between various grammatical functions, given that the lexical forms adopted by subjects and objects differ in Dutch. The rationale behind this methodology is that by systematically removing all lexical variations associated with third-person singular pronouns, the model will develop the capability to identify any token in this grammatical position as a pronoun, irrespective of its lexical form. This method has not previously been tested in a similar setup. \citet{lauscher-etal-2022-welcome}, who introduce this methodology, conduct tests in a reversed configuration, training a model on regular data and evaluating its performance on delexicalised data. In this context, the model did not exhibited a good performance. They additionally try both 
training and testing 
on delexicalised data, which results in a satisfactory performance. However, this experimental design does not faithfully simulate a scenario in which a model is debiased through delexicalised data, but 
deployed on naturally occurring data, which includes pronouns in their lexical forms. 

We evaluate both debiasing methods in two conditions: (i) \textbf{Fine-tuning the model from scratch} on the respective debiasing dataset, and (ii)  \textbf{continual fine-tuning} the original wl-coref model, initially trained on the regular \mbox{SoNaR-1} data, with the respective debiasing dataset. 
Given that continual fine-tuning is computationally less demanding compared to fine-tuning from scratch,\footnote{Here, we do not take the computational costs of fine-tuning the original wl-coref model into account, because we consider this an off-the-shelf model and we want to isolate the costs of debiasing an existing model. } it would be a preferable debiasing approach, provided it achieves a satisfactory performance.
For consistency, the same hyperparameters are employed as for the regular model. 
Debiased models fine-tuned from scratch are trained for 20 epochs, while the continual fine-tuned models are trained for 10 epochs. 
The models are evaluated on the \textit{pronoun-specific} test sets. The debiasing performance is measured as the difference between performance on gender-neutral pronouns by the debiased model and the regular wl-coref model, measured through the \textsc{lea} F1-score and the pronoun score.

\subsubsection{Results}
\label{sec:debiasing_results}

\begin{table}[t]
\centering
\small
\caption{Debiasing experiment: \textsc{lea} F1 and pronoun scores (\textsc{ps}) on the \textit{pronoun-specific} test sets after debiasing, as the average across five random seeds. While delexicalisation does not improve the results, CDA reduces the performance gap between gendered and gender-neutral pronouns.}
\label{tab:full-retrain-results}
\begin{tabular}{llrrrr}
\toprule
\textbf{Model} & \textbf{Metric} &  \begin{tabular}[c]{@{}c@{}} \textbf{Hij} \\ (masculine)\end{tabular}  & \begin{tabular}[c]{@{}c@{}} \textbf{Zij} \\(feminine)\end{tabular}  & \begin{tabular}[c]{@{}c@{}} \textbf{Hen} \\ (gender-neutral)\end{tabular} & \begin{tabular}[c]{@{}c@{}} \textbf{Die} \\ (gender-neutral)\end{tabular}\\ \midrule
\textit{Original model} & \textsc{lea} & 51.29   ($\sigma$=0.42)   & 50.77 ($\sigma$=0.37)    & 49.14 ($\sigma$=0.68)   & 48.36 ($\sigma$=0.44)   \\
& \textsc{ps} (\%) &  88.36   ($\sigma$=0.89)   & 86.65 ($\sigma$=1.23)             & 75.85 ($\sigma$=2.93)   & 57.49 ($\sigma$=6.55) \\
\hline
\multicolumn{6}{c}{Fine-tuning the wl-coref model from scratch} \\
\hline
\rowcolor[HTML]{EFEFEF}
\textit{Delexicalisation}  & \textsc{lea} &      53.04 ($\sigma$=0.70)        & 53.31 ($\sigma$=0.53)  &   50.67 ($\sigma$=0.79)  & 50.69 ($\sigma$=0.64)         \\
\rowcolor[HTML]{EFEFEF}
& \textsc{ps} (\%)&  76.50 ($\sigma$=4.56)        &  82.79 ($\sigma$=2.42)  &   71.55 ($\sigma$=4.94)  & 61.89 ($\sigma$=5.53)         \\
\textit{CDA}   & \textsc{lea} & 54.44 ($\sigma$=0.41)    &  54.47 ($\sigma$=0.49)        &  54.40 ($\sigma$=0.33)  & 54.33 ($\sigma$=0.41)  \\ 
& \textsc{ps} (\%) & 86.88 ($\sigma$=1.64)    &  89.08 ($\sigma$=0.93)        &  88.02 ($\sigma$=0.74)  & 89.37 ($\sigma$=0.57)  \\ 
\hline
\multicolumn{6}{c}{Continual fine-tuning the wl-coref model} \\
\hline
\rowcolor[HTML]{EFEFEF}
\textit{Delexicalisation}  & \textsc{lea}   &  53.74 ($\sigma$=0.78)        & 53.53 ($\sigma$=0.78)  &   50.51 ($\sigma$=1.05)  & 50.07 ($\sigma$=0.90)         \\
\rowcolor[HTML]{EFEFEF}
& \textsc{ps} (\%)&  89.29 ($\sigma$=1.17)        &  88.76 ($\sigma$=0.98)  &   72.91 ($\sigma$=2.80)  & 57.17 ($\sigma$=1.95)         \\
\textit{CDA}  & \textsc{lea} &  54.57 ($\sigma$=0.59)    &  54.48 ($\sigma$=0.63)        &  54.50 ($\sigma$=0.58)  & 54.36 ($\sigma$=0.59)  \\ 
& \textsc{ps} (\%) & 90.52 ($\sigma$=0.44)    & 90.60 ($\sigma$=0.33)        & 90.16 ($\sigma$=0.51)  & 89.60 ($\sigma$=0.50)  \\ 
\bottomrule
\end{tabular}
\end{table}

Table \ref{tab:full-retrain-results} displays 
the performance scores 
after (a) fine-tuning from scratch and (b) continual fine-tuning the original wl-coref model, using the two debiasing techniques. Here we discuss the main observations. 

\textbf{Delexicalisation does not successfully debias the model}. The pronoun scores remain low in both conditions, and even deteriorate in the continual fine-tuning condition. This suggests that the removal of lexical information alone is insufficient to effectively 
improve the model's 
performance on gender-neutral pronouns. 

\textbf{The application of CDA fine-tuning from scratch shows substantial improvements on gender-neutral pronouns.} The pronoun scores exceed 86\% for all pronouns, representing an improvement of 31.88 percentage points for \textit{die} and 12.17 percentage points for \textit{hen}.
Furthermore, the fine-tuned from scratch model sustains a high performance for \textit{hij} and \textit{zij}, despite not encountering these pronouns during training. This implies that the base model's pre-training already imparts sufficient familiarity with these pronouns.\footnote{The performance for \textit{zij} surpasses that of \textit{hij} for both models, possibly due to the continued occurrence of \textit{zij} in the corpus as a third-person plural pronoun, while \textit{hij} ceases to appear altogether, lacking an alternative meaning in Dutch.}

\textbf{Continual fine-tuning with CDA results in the best debiasing outcomes}. The F1-scores across pronoun-specific test sets surpass 54.0, an improvement for all pronouns. Moreover, all pronoun scores exceed 89.5\%, achieving even slightly higher scores than the fine-tuned from scratch CDA model. These results are encouraging, particularly considering that continual fine-tuning already was the preferred method, due to its computational efficiency.

Lastly, we evaluate the impact of debiasing on the performance on the original test set in Appendix \ref{app:debiasing_impact}, to investigate if any knowledge is lost through the debiasing process. Table \ref{tab:debiased_regular_performances} shows that all debiased models exhibit a small performance drop in comparison to the original model, while this decline is most pronounced for the delexicalised models. The decrease is smaller for the CDA models, with a drop of only -0.58 in the continual fine-tuning setting.


\subsection{Low-resource debiasing experiment}
\label{sec:debiasing_data_amount}
We now investigate whether the best debiasing method can also be effectively applied in a scenario with limited data, because large corpora for debiasing may not always be available. Moreover, debiasing with a smaller corpus reduces the computational costs. We use \textit{CDA through continual fine-tuning}, because this method obtains the best debiasing results (Section~\ref{sec:debiasing-experiments}), reducing the performance gap between gendered and gender-neutral pronouns to less than 1\%. 

We implement CDA with the same experimental setup as before, but we only use fractions of the \textit{gender-neutral} training set, specifically 10\%, 5\%, 2.5\%, and 1.25\%, corresponding to 62, 30, 15, and 7 documents respectively.\footnote{It is important to note that in the \textit{gender-neutral} training set, the usage of the pronouns \textit{hen} and \textit{die} alternates between documents, with each pronoun featured in only 50\% of the documents. Thus, when debiasing with, for instance, 30 documents, each pronoun is present in only fifteen documents.} Five partitions are used for each training size fraction, of which the average scores are reported. In the interest of computational efficiency, we only use one seed. 

\begin{table}[]
\centering
\small
\caption{Low-resource debiasing experiment: Pronoun scores after applying \textit{CDA continual fine-tuning} with various fractions of the full \textit{gender-neutral} training set. The reported scores represent average across five data partitions. The performances already improve after debiasing with a few documents. }
\label{tab:data_size_results}
\begin{tabular}{rrllll}
\toprule
\textbf{Percentage} & \textbf{\begin{tabular}[r]{@{}l@{}}\# Train\\docs\end{tabular}} & \begin{tabular}[c]{@{}l@{}} \textbf{Hij} \\ (masculine)\end{tabular}  & \begin{tabular}[c]{@{}l@{}} \textbf{Zij} \\(feminine)\end{tabular}  & \begin{tabular}[c]{@{}l@{}} \textbf{Hen} \\ (gender-neutral)\end{tabular} & \begin{tabular}[c]{@{}l@{}} \textbf{Die} \\ (gender-neutral)\end{tabular}\\ \midrule
\textit{100\%}  	& 625 & 90.76\%    & 90.60\%        & 89.94\%   & 89.67\%  \\
\hline
\textit{10\%}    & 62  &  92.41\% ($\sigma$=0.19)    & 91.26\% ($\sigma$=0.41)        & 88.64\% ($\sigma$=0.79)  & 85.42\% ($\sigma$=0.94)       \\
\textit{5\%}    &  30   &   92.02\% ($\sigma$=0.48)    & 90.66\% ($\sigma$=0.43)        & 87.32\% ($\sigma$=0.90)  & 83.65\% ($\sigma$=1.08)   \\ 
\textit{2.5\%}    & 15    & 91.40\% ($\sigma$=0.64)    & 89.96\% ($\sigma$=0.54)        & 85.09\% ($\sigma$=0.89)        & 79.48\% ($\sigma$=0.94)  \\ 
\textit{1.25\%}    & 7    & 91.36\% ($\sigma$=0.62)    & 90.25\% ($\sigma$=0.58)        & 85.12\% ($\sigma$=1.06)  & 78.44\% ($\sigma$=1.81)  \\ 
\hline
\textit{Original model }   & 0    & 88.19\% & 	86.66\% & 	78.79\% &	65.77\%   \\ 
\bottomrule
\end{tabular}
\end{table}

\subsubsection{Results}
Table \ref{tab:data_size_results} presents the outcomes. \textbf{The pronoun scores improve after debiasing with just a few debiasing documents.} With a tiny dataset of 7 documents (3--4 documents per pronoun), substantial improvements (+12.67 percentage points for \textit{die}; +6.33 percentage points for \textit{hen}) are observed. Furthermore, by using 5\% of the documents (15 documents per pronoun), the pronoun scores already surpass 80\% for the gender-neutral pronouns. This is in line with \citet{bjorklund2023computer}, who observe that including gender-neutral pronouns in 2\% of the training instances leads to a satisfying POS-tagging performance on these pronouns.\footnote{Because two pronouns are simultaneously debiased in the current study, the 5\% setting here corresponds to \citeauthor{bjorklund2023computer}'s 2\% setting, as the debiasing documents alternate between the usage of \textit{hen} (2.5\%) and \textit{die} (2.5\%).} The gap between debiasing with 5\% of the documents and all documents is only 2.62 percentage points for \textit{hen} and 6.02 percentage points for \textit{die}. These results show that effective debiasing can be achieved with reduced access to resources.

\subsection{Unseen pronouns experiment}
\label{sec:unseen_pronouns_exp}

\subsubsection{Setup}
This experiment evaluates the ability of all models to process pronouns that have not previously been encountered by the model. The reason for this evaluation is that novel (neo-)pronouns may be popularised in the future \cite{lauscher-etal-2022-welcome}. Creating systems that will correctly process these pronouns or can easily adapt is preferred over debiasing strategies that are tailored exclusively to specific pronouns, as the latter require recurrent debiasing efforts each time a new pronoun gains popularity. 


We evaluate the original and the debiased models on the \textit{unseen test set}: a version of the dataset wherein all third-person pronouns are substituted by a neopronoun $p$, randomly selected from a set of six Dutch neopronouns: $p \in$~$\{$ \textit{dee/dem/dijr, dij/dem/dijr, nij/ner/nijr, vij/vijn/vijns, zhij/zhaar/zhaar, zem/zeer/zeer} $\}$.\footnote{Pronouns were extracted from the list on \url{https://nl.pronouns.page/voornaamwoorden}} 
Prior studies on debiasing coreference resolution systems \cite{lauscher-etal-2022-welcome, zhao-etal-2018-gender, zhao-etal-2019-gender} have not included evaluations that focus on previously unseen pronouns. 

We do not expect our previous debiased models to improve the performance on neopronouns: 
although the delexicalisation method was specifically designed to enable the model to process pronouns of diverse lexical forms, 
its debiasing performance showed unsatisfactory for gender-neutral pronouns (see Section \ref{sec:debiasing-experiments}). Consequently, we expect this method will similarly fall short on 
debiasing unseen pronouns. Moreover, CDA relies on instructing the model to process a particular pronoun by exposing it directly to the pronoun's lexical form. Consequently, it is also expected that CDA will not enhance performance on unseen pronouns, as the model lacks exposure to these specific pronouns. 


\subsubsection{Results}
\label{sec:unseen_exp_results}
Table \ref{tab:expp_4_results} reports the results for this experiment. The original model has an unsatisfactory performance on unseen pronouns. Moreover, consistent with our expectations, \textbf{neither of the debiasing methods improves the performance on unseen pronouns:}  
the highest scores are for the \textit{continual fine-tuned CDA} model, which achieves a pronoun score of 36.3 percent points lower than on \textit{die} pronouns (Table \ref{tab:full-retrain-results}).
%

\begin{table}[t]
\centering
\small
\caption{Unseen pronouns experiment: Model performances, in terms of the \textsc{lea} metric and the pronoun score, on the \textit{unseen} test set. This test set includes neopronouns that were not previously encountered by the models. The models include the original wl-coref model, alongside four models that were debiased through fine-tuning from scratch or continual fine-tuning, using delexicalisation or CDA. The reported scores are the average of five random seeds. All models perform poorly on previously unseen pronouns.}
\label{tab:expp_4_results}
\begin{tabular}{llll|l}
\toprule
                        & \multicolumn{1}{c}{\textbf{Precision}} & \multicolumn{1}{c}{\textbf{Recall}} & \multicolumn{1}{c}{\textbf{F1}} & \textbf{Pronoun score} \\
\midrule
\textit{Original model} & 52.83 ($\sigma$=2.35)             & 44.37 ($\sigma$=2.92)           & 48.12 ($\sigma$=0.66)       & 46.68\%  ($\sigma$=2.31)                            \\
\hline
\multicolumn{5}{c}{Fine-tuning the wl-coref model from scratch} \\
\hline
\textit{Delexicalisation}               &   52.17 ($\sigma$=1.94)             & 49.55 ($\sigma$=1.95)           & 50.77 ($\sigma$=0.46)       & 48.03\%  ($\sigma$=2.01)                            \\
\textit{CDA}               &   53.46 ($\sigma$=2.56)             & 49.66 ($\sigma$=3.23)           & 51.36 ($\sigma$=0.61)       & 51.72\%  ($\sigma$=2.90)                            \\
\hline
\multicolumn{5}{c}{Continual fine-tuning the wl-coref model} \\
\hline
\textit{Delexicalisation}               & 50.98 ($\sigma$=0.83)             & 51.03 ($\sigma$=1.66)           & 50.99 ($\sigma$=0.72) &  49.56\%  ($\sigma$=2.07)                            \\
\textit{CDA}               & 53.01 ($\sigma$=0.73)             & 50.22 ($\sigma$=1.05)           & 51.57 ($\sigma$=0.61)       & 53.37\%  ($\sigma$=3.55)                            \\
\bottomrule
\end{tabular}

\end{table}

\begin{figure}[t]
    \centering
    \begin{minipage}[t]{.48\textwidth}
        \centering
        \includegraphics[scale=0.38]{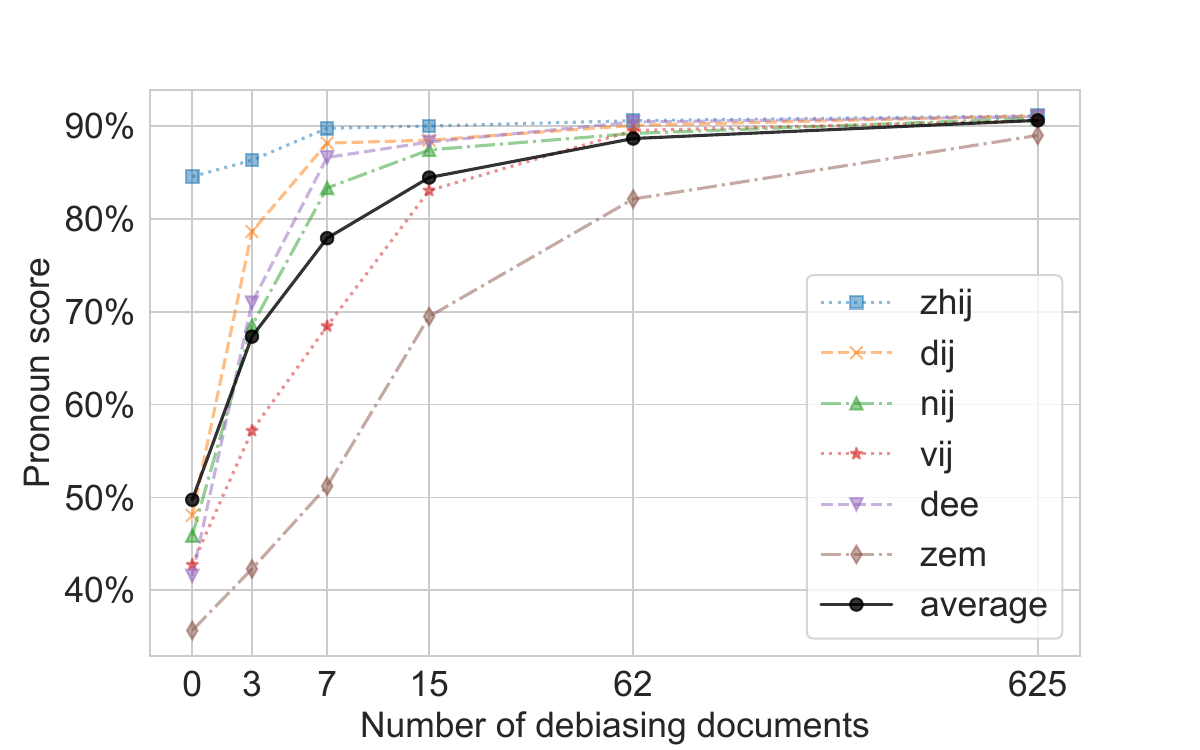}
    \caption{Neopronouns debiasing experiment: Pronoun scores across six neopronouns as a function of the number of debiasing documents included in \textit{CDA continual fine-tuning}. The black dotted line indicates the average pronoun scores across the different neopronouns. Reported scores are the average of five data partitions.}
    \label{fig:neo_debiasing}
    \Description[The average pronoun scores for neopronouns improve, even with just a few debiasing documents.]{The average pronoun scores for neopronouns improve, even with just a few (3/7/15) debiasing documents.}
    \end{minipage} \hspace{0.2cm}%
    \begin{minipage}[t]{0.48\textwidth}
        \centering
        \includegraphics[scale=0.38]{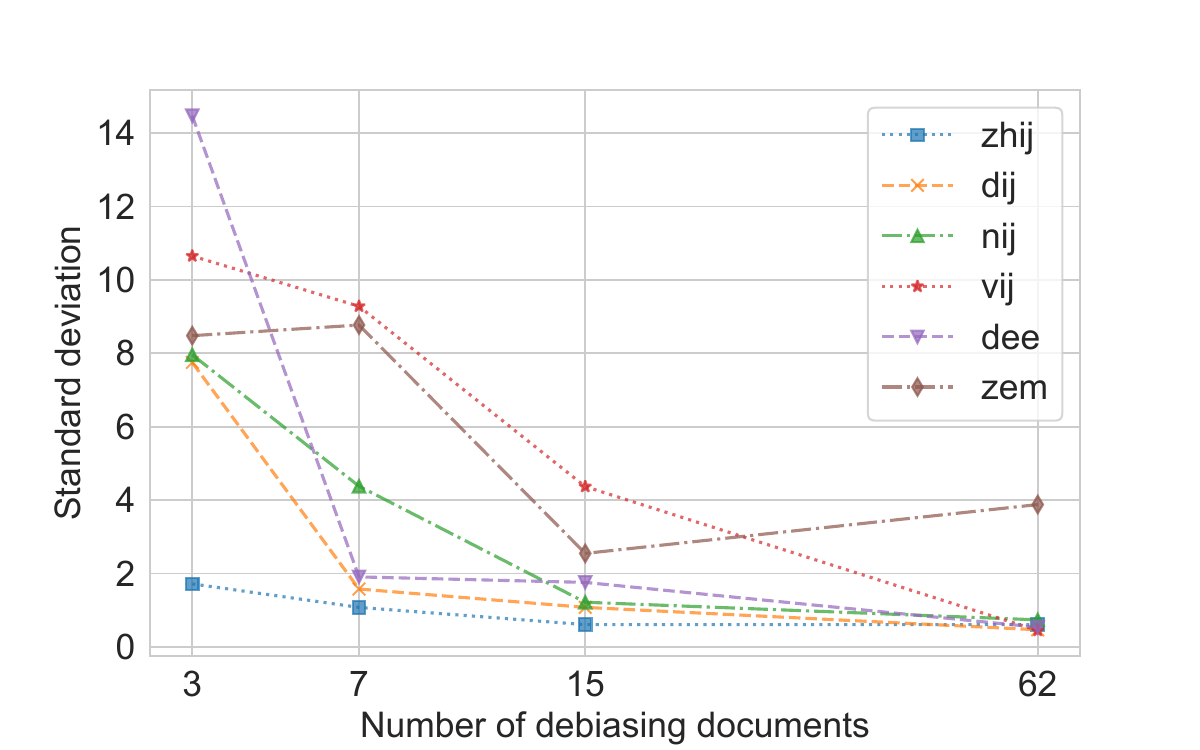}
    \caption{Neopronouns debiasing experiment: Standard deviations of the pronoun scores across five different data partitions, for six neopronouns, as a function of the number of debiasing documents included in \textit{CDA continual fine-tuning}. The zero and 625 training documents settings are excluded from this figure, because these two settings only use one data partition (an empty set or the full set).}
    \Description[The standard deviations are high with a low number of debiasing documents, but decrease when more documents are used.]{The standard deviations are high with a low number of debiasing documents, but decrease when more documents are used.}
    \label{fig:neo_debiasing_std}
    \end{minipage}
\end{figure}

\subsection{Neopronouns debiasing experiment}
\label{sec:neo_debiasing}
In light of the poor performance on processing previously unseen pronouns, we now assess whether continual CDA fine-tuning, the most successful method in our debiasing experiment, is able to improve the processing of neopronouns. We adopt a low-resource setting for two reasons: (1) it is computationally desirable as it requires minimal resources and computational costs; (2) 
the experiments in Section \ref{sec:debiasing_data_amount} showed satisfactory debiasing results in such a setup. 
Debiasing for neopronouns is different from debiasing gender-neutral pronouns, because the former types do not yet exist in the language at all. This distinction introduces potential advantages and challenges to the debiasing process: 
a higher amount of debiasing data may be required to familiarise the model with these new types, but on the other hand, the absence of pre-existing usage patterns may alleviate ambiguity.\footnote{For example, the gender-neutral pronoun \textit{hen} may exhibit ambiguity in certain sentence structures as it can refer to either third-person singular or plural. In contrast, a neopronoun consistently maintains a singular and unambiguous referent, potentially facilitating the debiasing process.}

\subsubsection{Setup}
The setup is similar to that of the low-resource debiasing experiment 
(Section \ref{sec:debiasing_data_amount}). However, here, we debias each pronoun individually, to allow a comparison between neopronouns. For each pronoun, we create a debiasing and a test set with this particular pronoun inserted. 
Similarly, we vary the amount of debiasing documents between 3~ ($0.625\%$), 7 ($1.25\%$), 15 ($2.5\%$) and 62 ($10\%$) documents, using five data partitions. 
For comparison, we include the model performance 
(a) before debiasing and (b) after debiasing with the full training set (625 documents) 
as baselines.

\subsubsection{Results}
The results of the debiasing process, in terms of pronoun scores, are presented in Figure~\ref{fig:neo_debiasing}. 

\textbf{The performance on the different pronouns before debiasing varies a lot.} 
For instance, \textit{zhij}, which is formed by combining the known gendered pronouns \textit{hij} and \textit{zij}, demonstrates an impressive initial performance of 84.55\%. In contrast, \textit{zem}, the neopronoun that least resembles known pronouns, exhibits the lowest initial score of 35.66\%.

\textbf{A small number of debiasing documents can improve the performance.}
We observe a consistent debiasing trend across the pronouns.
For example, using only three training documents results in a substantial average performance improvement from 49.7\% to 67.3\% (+17.6 percentage points), 
despite high standard deviations (see Figure \ref{fig:neo_debiasing_std}). 
A high standard deviation is sensible in this setting, considering the significant variation in length and pronoun frequency across training documents. 
Moreover, as the number of debiasing documents increases, 
the performances keep improving and standard deviations generally decrease. 
With the inclusion of 15 debiasing documents, the average performance on neopronouns already improves to 84.5\%. 
It is worth noting that for certain pronouns (\textit{zhij, dij, nij}), 
performance begins to plateau with additional debiasing documents, 
indicating a saturation of debiasing data; 
whereas others (particularly \textit{zem}) continues to benefit from further debiasing. Additionally, we discuss the effect of the tokeniser in Appendix \ref{app:tokenisation}. 

Taken together, the outcomes of this experiment are promising.
Satisfactory results are achieved across various sets of neopronouns using CDA continual fine-tuning with a limited dataset. These findings demonstrate the feasibility of future-proof gender-inclusive debiasing with minimal resource requirements and low computational costs.

\section{Discussion and conclusion}
\label{sec:discussion}
The results of this study have two main implications. First, the debiasing results show that applying CDA manifests a considerable improvement, even when applied through continual fine-tuning with just a handful of documents. 
This outcome aligns with the finding of \citet{bjorklund2023computer} that effective debiasing of gender-neutral pronouns can be achieved with a low number of debiasing instances; and more generally it underscores the feasibility of debiasing in non-binary contexts with minimal resources and low computational costs. This result is particularly noteworthy given the absence of gender-neutral pronouns in the original training data and the general novelty of gender-neutral pronouns in the Dutch language. 

Second, the results observed in this study suggest that there exists an opportunity for NLP technologies to be at the forefront of emancipation movements, by enabling systems to adeptly process emerging languages structures, which are embraced by pioneers but are not yet prevalent throughout broader societies. The Dutch gender-neutral pronouns and neopronouns serve as illustrative instances of such emergent linguistic constructs. The implementation of NLP technologies in this context holds the potential to facilitate the wider adoption of these innovative structures within societies, by showing people an example of how to correctly use these structures.

A limitation of this study is that we only consider a single model in our evaluation. Future research could extend the scope by exploring potential trends across various models. Moreover, future research could assess the applicability of the current setup to other languages. For example, for languages like Italian or French, in which gender is more intricately woven into grammatical structures, the debiasing task may be more complex.

\section{Ethical considerations}
 We zoom in on gender-neutral pronouns alone, and discard any other dimension in which the language of non-binary individuals may differ from that of people with a binary gender identity, such as vocabulary\footnote{For example, non-binary individuals may use \textit{neo}nouns such as \textit{brus} (\textit{sibling}): a contraction of \textit{broer} (\textit{brother}) and \textit{zus} (\textit{sister}).} and style. Despite the fact that we observe that debiasing through CDA improves the performance on gender-neutral pronouns, this does not imply that the performance of the coreference resolution system would also improve on real-world data from non-binary individuals, because the data considered in this study still stems from binary-gendered contexts. Therefore, an important direction for future work would be to test, and if necessary debias, model performance on Dutch data from transgender individuals, for instance through creating a Dutch equivalent of the GICoref corpus \cite{cao-daume-iii-2020-toward}.

Moreover, this study has not actively involved non-binary and transgender individuals in the designing, debiasing and evaluation process, and instead was for the main part conducted by cisgender individuals in a binary gendered environment. We recognise that this may have led to overlooking important barriers, risks or opportunities relating to the emancipation of non-binary individuals. We hope that the current study, which exclusively looks at gender-neutral pronouns, 
can be considered a small step, at the beginning stages of achieving emancipation and a fair treatment of non-binary individuals in Dutch language technologies. But, in the steps that follow towards achieving this goal, the active involvement of non-binary individuals, for instance through participatory design initiatives \cite{caselli-etal-2021-guiding}, is essential \cite{devinney2022theories}.

\begin{acks}
Dong Nguyen is funded by the Veni research programme
with project number VI.Veni.192.130, which is
(partly) financed by the Dutch Research Council
(NWO). We want to thank Antal van den Bosch, Guusje Juijn, Tim Koornstra, Michael Pieke, Daan van der Weijden, Shiyi Butter and Silin Chen for their constructive feedback. We thank the anonymous reviewers for their efforts in reviewing this paper and their interesting questions.
\end{acks}

\bibliographystyle{ACM-Reference-Format}
\bibliography{sample-base, anthology}

\appendix

\clearpage
\section{Gendered nouns rewriting rules}
\label{sec:app_rewriting_rules}

\begin{table}[h!]
\centering
\small
\caption{Rewriting rules for gendered Dutch nouns to a gender-neutral version of this word (part 1). Not all Dutch words have a gender-neutral alternative however. * marks difficult cases, for which some meaning is lost.}
\label{tab:rewriting_rules1}
\begin{tabular}{ll}
\toprule
 \textbf{Gendered noun} & \textbf{Gender-neutral noun} \\
\midrule
         tante &          familielid* \\
           oom &          familielid* \\
        jongen &                kind \\
        meisje &                kind \\
           man &             persoon \\
         vrouw &             persoon \\
        mannen &            personen \\
       vrouwen &            personen \\
         broer &          familielid* \\
           zus &          familielid* \\
      broertje &          familielid* \\
         zusje &          familielid* \\
     broertjes &        familieleden* \\
        zusjes &        familieleden* \\
        broers &        familieleden* \\
        zussen &        familieleden* \\
          meid &             persoon \\
         vader &               ouder \\
        moeder &              ouder \\
        vaders &              ouders \\
       moeders &              ouders \\
          zoon &                kind \\
         zonen &            kinderen \\
       dochter &                kind \\
      dochters &            kinderen \\
         nicht &          familielid* \\
       nichtje &          familielid* \\
      nichtjes &        familieleden* \\
       nichten &        familieleden* \\
          neef &          familielid* \\
        neefje &          familielid* \\
       neefjes &        familieleden* \\
  kleindochter &           kleinkind \\
     kleinzoon &           kleinkind \\
 kleindochters &       kleinkinderen \\
    kleinzonen &       kleinkinderen \\
           oma &          grootouder \\
           opa &          grootouder \\
   grootmoeder &          grootouder \\
    grootvader &          grootouder \\
          dame &             persoon \\
          heer &             persoon \\
         dames &            personen \\
         heren &            personen \\
        koning &         staatshoofd \\
      koningin &         staatshoofd \\

\bottomrule
\end{tabular}
\end{table}

\begin{table}[t]
\centering
\small
\caption{Rewriting rules for gendered Dutch nouns to a gender-neutral version of this word (part 2). Not all Dutch words have a gender-neutral alternative however. * marks difficult cases, for which some meaning is lost.}
\label{tab:rewriting_rules2}
\begin{tabular}{ll}
\toprule
 \textbf{Gendered noun} & \textbf{Gender-neutral noun} \\
\midrule
      koningen &       staatshoofden \\
   koninginnen &       staatshoofden \\
       mevrouw &             persoon* \\
        meneer &             persoon* \\
     jongedame &             jongere* \\
      jongeman &             jongere* \\
    politieman &        politieagent \\
  politievrouw &        politieagent \\
    brandweerman &       brandweermens \\
brandweervrouw &       brandweermens \\
       prinses &               edele* \\
         prins &               edele* \\
    prinsessen &              edelen* \\
       prinsen &              edelen* \\
    kroonprins &       troonopvolger \\
  kroonprinses &       troonopvolger \\
     schrijver &              auteur \\
   schrijfster &              auteur \\
           juf &          leerkracht \\
       meester &          leerkracht \\
        leraar &          leerkracht \\
       lerares &          leerkracht \\
         bruid &         jonggehuwde \\
     bruidegom &         jonggehuwde \\
      tovenaar &             magiër \\
          heks &             magiër \\
    stiefvader &          stiefouder \\
   stiefmoeder &          stiefouder \\
     stiefzoon &           stiefkind \\
  stiefdochter &           stiefkind \\
        weduwe &         nabestaande* \\
     weduwnaar &         nabestaande* \\
           kok &                chef \\
        kokkin &                chef \\
    kunstenaar &             artiest \\
  kunstenaares &             artiest \\
        vriend &                maat* \\
      vriendin &                maat* \\
      vriendje &             partner* \\
 vriendinnetje &             partner* \\
\bottomrule
\end{tabular}
\end{table}

\clearpage
\section{Hyperparameter tuning}
\label{app:hyperparams}
The results of the hyperparameter search for the learning rate is present in Table \ref{tab:learning-rate} and the corresponding results for the BERT learning rate can be found in Table \ref{tab:bert-learning-rate}.

\begin{table}[t]
\centering
\small
\caption{Learning rate tuning of the wl-coref model, using XLM-RoBERTa \cite{conneau-etal-2020-unsupervised} as its base model. \textsc{lea} performance scores on the SoNaR-1 development set. Models were trained for twenty epochs, keeping all other hyperparameters the same as \citet{dobrovolskii-2021-word}. The best F1-score is found for a learning rate of $5e^{-4}$.}
\label{tab:learning-rate}
\begin{tabular}{lllllll}
\toprule
\textbf{Learning rate} & \textbf{Precision}    & \textbf{Recall}     & \textbf{F1}     \\
\midrule
\rowcolor[HTML]{EFEFEF} 
$8e^{-4}$                     & 52.9           & 56.4        & 54.6       \\
$6e^{-4}$                       & 50.6           & \textbf{59.1 } & 54.5   \\
\rowcolor[HTML]{EFEFEF} 
$5e^{-4}$               & 51.2           & 58.6     & \textbf{54.7 }        \\
$4e^{-4}$                  & 52.8           & 56.0       & 54.3        \\
\rowcolor[HTML]{EFEFEF} 
$3e^{-4}$ (original value)                        &52.6           &54.7       &53.6            \\
$1e^{-4}$                    &54.3           &49.8       &52.0       \\
\rowcolor[HTML]{EFEFEF} 
$5e^{-5}$                    &58.2           &37.7        &45.8       \\
$4e^{-5}$               &58.4           &35.1        &43.8          \\
\rowcolor[HTML]{EFEFEF} 
$3e^{-5}$                  &60.1           &24.8        &35.1      \\
$2e^{-5}$                   &60.0          &18.7      &28.5           \\
\rowcolor[HTML]{EFEFEF} 
$1e^{-5}$                           & \textbf{63.9 } & 5.7      &10.5       \\
$5e^{-6}$                          &52.8           &9.9        &16.7     \\
\bottomrule
\end{tabular}
\end{table}

\begin{table}[]
\centering
\small
\caption{BERT learning rate tuning of the wl-coref model, using XLM-RoBERTa \cite{conneau-etal-2020-unsupervised} as its base model. \textsc{lea} performance scores on the SoNaR-1 development set. Models were trained for twenty epochs, using a learning rate  $= 5e^{-4}$, and keeping all other hyperparameters the same as \citet{dobrovolskii-2021-word}. The best F1-score is found for a BERT learning rate of $3^{-5}$.}
\label{tab:bert-learning-rate}
\begin{tabular}{lcccccc}
\toprule
\multicolumn{1}{l}{\textbf{BERT learning rate}} &  \multicolumn{1}{l}{\textbf{Precision}} & \multicolumn{1}{l}{\textbf{Recall}} & \multicolumn{1}{l}{\textbf{F1}} \\
\rowcolor[HTML]{EFEFEF} 
\midrule
$1e^{-4}$                                                                                &50.2                           &58.9          &54.2                  \\
$1e^{-5}$ (original value)             & 51.2           & 58.6     & 54.7         \\ \rowcolor[HTML]{EFEFEF} 
$2e^{-5}$                                                                    &53.2                           &57.3     &55.2                      \\

$3e^{-5}$                                                                          &52.0                          & \textbf{59.5 }    & \textbf{55.5 }             \\
\rowcolor[HTML]{EFEFEF} 
$4e^{-5}$                                                                              &52.0                          &57.9      &54.8                         \\
$5e^{-5}$                                                                                        & \textbf{54.1 }                 &56.6       &55.3                     \\
\rowcolor[HTML]{EFEFEF} 
$5e^{-6}$                                                         &   51.2                           &56.6       &53.8     \\
\bottomrule
\end{tabular}
\end{table}

\section{Model comparison}
\label{app:models}
Compared to the neural e2e-Dutch model,\footnote{\url{https://github.com/Filter-Bubble/e2e-Dutch}} wl-coref's test F1-score is $6$ points lower. However, improvements over e2e-Dutch are (1) a lower complexity \cite{dobrovolskii-2021-word} and a smaller difference between the development and test performance (F1~$\Delta = -3.7$ for e2e-Dutch and  F1 $\Delta = +0.3$ for wl-coref), suggesting minimal overfitting. Moreover, the wl-coref model notably outperforms the rule-based dutchcoref model \cite{van2019dutch} ($+11.6$ in test F1-score). 

\section{Debiasing impact on general model performance}
\label{app:debiasing_impact}
We evaluate the impact of debiasing on the model's performance on the original test set, in order to inspect whether any knowledge
is lost through the debiasing process. We compare the two debiasing techniques (see Section \ref{sec:debiasing-experiments}). Table \ref{tab:debiased_regular_performances} shows that applying CDA results in a smaller performance drop than delexicalisation.

\begin{table}[]
\centering
\small
\caption{Debiasing experiment: Average \textsc{lea} F1-scores achieved by the debiased models on the regular SoNaR-1 test data, as the average across five random seeds. This evaluation aims to assess potential losses in abilities through the debiasing process. CDA leads to a smaller drop in performance than delexicalisation.}
\label{tab:debiased_regular_performances}
\begin{tabular}{lcc}
\toprule
\textbf{Model}          & \multicolumn{1}{l}{\textbf{F1 performance regular test set}} & \multicolumn{1}{l}{\textbf{$\Delta$ original model}} \\
\midrule
\textit{Original model} & 55.57 ($\sigma$ = 0.46)                         & -                                                                 \\
\hline
\textit{Delexicalisation full}     & 53.04 ($\sigma$ = 0.58)                          & -2.53                                                               \\
\textit{Delexicalisation fine}     & 54.38 ($\sigma$ = 0.86)                         & -1.37                                                             \\
\textit{CDA full}       & 54.48 ($\sigma$ = 0.51)                         & -1.27                                                             \\
\textit{CDA fine}       & 55.17 ($\sigma$ = 0.47)                         & -0.58    \\
\bottomrule
\end{tabular}
\end{table}

\section{Tokenisation}
\label{app:tokenisation}
We directly use the XLM-RoBERTa-Base sentence piece tokeniser, which is trained on multilingual pre-training data. The tokeniser represents both gender-specific (\textit{hij}, \textit{zij}) and gender-neutral pronouns (\textit{hen, die}) as single tokens, suggesting that their performance disparities are not a result from the tokenisation. However, for neo-pronouns, debiasing efficiency seems to correlate with token quantity. In Figure \ref{fig:neo_debiasing}, the neopronouns \textit{zhij/dij/dee/nij/vij/zem} (ordered by their debiasing efficiency) are represented with 3/2/2/2/1/1 tokens. One notable comparison is between \textit{vij} and \textit{dee}: they have similar performance before debiasing, but after debiasing \textit{dee} (2 tokens) outperforms \textit{vij} (1 token) by a substantial margin. 
\end{document}